\documentclass[11pt]{article}

\usepackage[preprint]{acl}

\usepackage{times}
\usepackage{latexsym}

\usepackage{tikz}
\usetikzlibrary{arrows.meta, positioning}

\usepackage[T1]{fontenc}
\usepackage[utf8]{inputenc}

\usepackage{microtype}

\usepackage{inconsolata}
\usepackage{todonotes}

\usepackage{graphicx}
\usepackage{tabularx}
\usepackage{multirow}
\usepackage{verbatim}
\usepackage{amssymb}
\usepackage{enumitem}
\usepackage{amsmath}
\usepackage[most]{tcolorbox}
\usepackage{booktabs}  
\usepackage{makecell}   % for \makecell
\usepackage{array}      % for \arraybackslash in column definitions
\usepackage{subcaption}
\usepackage{graphicx}      % Required for \adjustbox (often a dependency)
\usepackage{adjustbox}     % For \adjustbox to scale the table
\usepackage{xcolor} % For \rowcolor and coloring in tables
\usepackage{arydshln}   % For \hdashline (horizontal dashed line)
\usepackage{amsmath}

\title{LingVarBench: Benchmarking LLMs on Entity Recognitions and Linguistic Verbalization Patterns in Phone-Call Transcripts}

\author {
    Seyedali Mohammadi\thanks{These authors contributed equally.},
    Manas Paldhe\footnotemark[1],
    Amit Chhabra,
    Youngseo Son,
    Vishal Seshagiri\\
    Infinitus Systems, Inc.,
    San Francisco, CA, USA\\
    \{ali.mohammadi, manas.paldhe, amit.chhabra, youngseo.son, vishal.seshagiri\}@infinitus.ai
}

\begin{document}
\maketitle

\begin{abstract}
We study structured entity extraction from phone-call transcripts in customer-support and healthcare settings, where annotation is costly and data access is limited by privacy and consent. Existing methods degrade under disfluencies, interruptions, and speaker overlap, yet large real-call corpora are rarely shareable. We introduce \textsc{LingVarBench}, a benchmark and semantic synthetic data generation pipeline that generates linguistically varied training data via (1) LLM-sampled entity values, (2) curated linguistic verbalization patterns covering diverse disfluencies and entity-specific readout styles, and (3) a value–transcript consistency filter. Using this dataset, DSPy’s SIMBA automatically synthesizes and optimizes extraction prompts, reducing manual prompt engineering and targeting robustness to verbal variation. On real customer transcripts, prompts optimized solely on \textsc{LingVarBench} outperform zero-shot baselines and match or closely approach human-tuned prompts for structured entities such as ZIP code, date of birth, and name (F1 $\approx 94\text{--}95$). For subjective questionnaire items, optimized prompts substantially improve over zero-shot performance and approach human-tuned prompts. \textsc{LingVarBench} offers a practical and cost-efficient path to deployment in a direct-answer setting, with real annotations later enabling additional refinement.
\end{abstract}

\section{Introduction}

% Motivation
% \begin{figure}[t]
% \centering
% \begin{tikzpicture}[node distance=1.1cm, font=\small]
% \tikzset{
%   box/.style={rectangle, rounded corners, draw=black, thick, align=center, minimum width=0.9\linewidth, minimum height=1cm},
%   arrow/.style={-{Stealth[length=2.5mm]}, thick}
% }
\begin{figure}[t]
    \centering
    \includegraphics[width=0.99\linewidth]{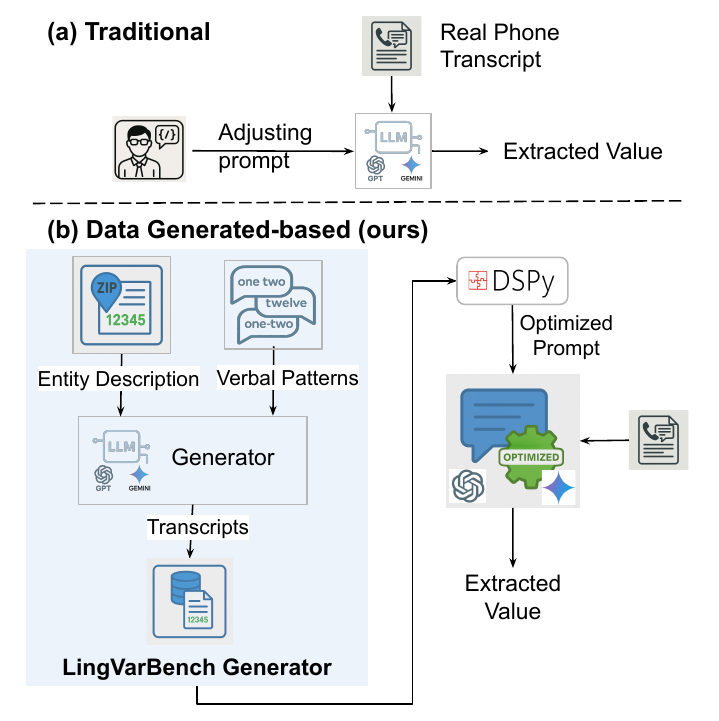}
    \caption{\footnotesize\textbf{Motivation for \textsc{LingVarBench}:} (a) Traditional entity extraction starts with little or no usable transcript data, leading to poor performance and requiring manual prompt tuning as transcripts arrive. (b) \textsc{LingVarBench} instead synthesizes diverse transcript verbalization patterns and uses DSPy + SIMBA to optimize prompts and generate robust, scalable evaluation data.}
    \label{fig:motivation}
\end{figure}

\begin{figure*}
  \centering
  \includegraphics[width=\linewidth]{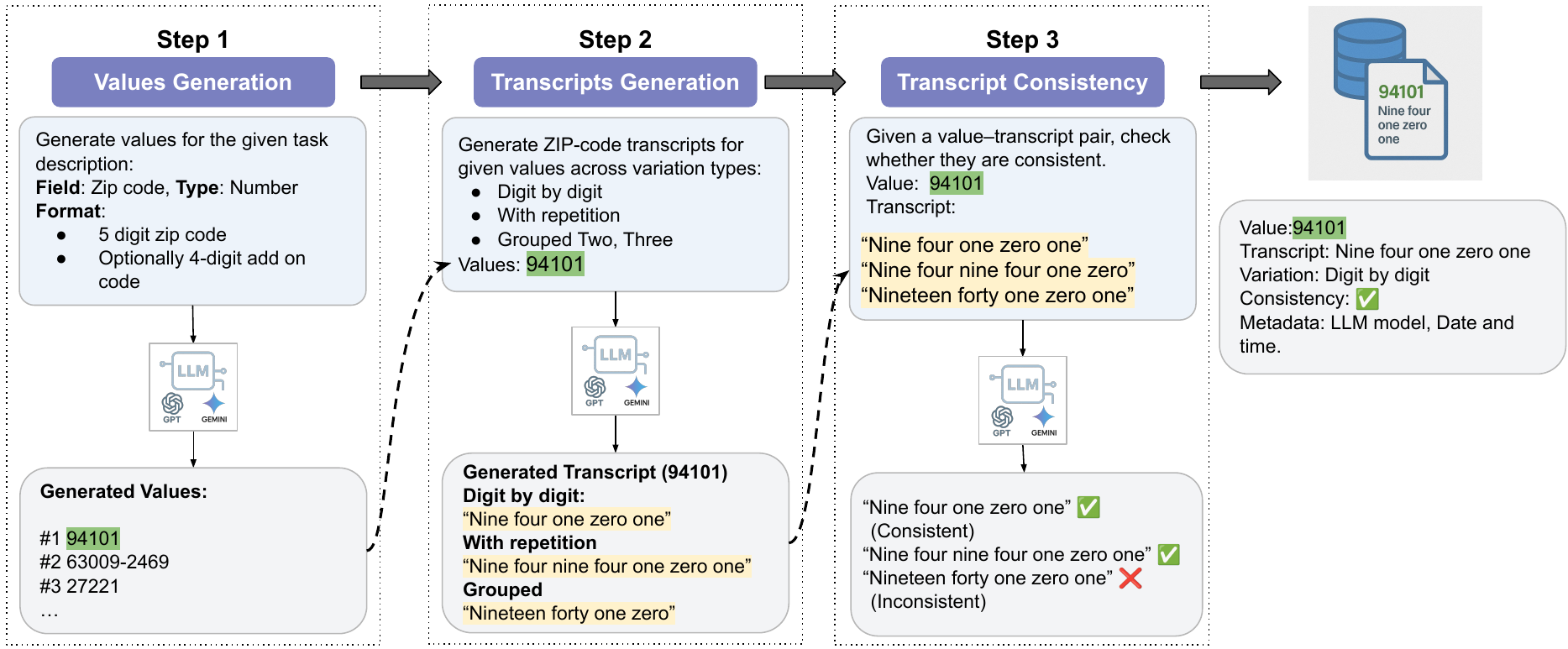}
  \caption{\footnotesize\textbf{Overview of the \textsc{LingVarBench} Synthetic Data Generation Framework:} A three-step pipeline for generating linguistically varied transcripts of structured fields. Canonical values (e.g., zip code) are transformed into diverse spoken-style utterances with consistency checks to ensure accurate recovery.}
  \label{fig:BenchFramework}
\end{figure*}

Voice-enabled AI is rapidly entering healthcare, powering clinical documentation, patient interaction, and administrative automation \citep{grandview_2024_healthcare_voice_ai,augnito_2024_voice_healthcare}. A core building block of these systems is \emph{structured information extraction} from patient--provider conversations: reliably recovering name, date of birth (DOB), ZIP code, and other fields from spontaneous speech \citep{foley_2024_hipaa_ai}. Unlike form-based EHR interfaces, voice AI must handle highly variable verbalizations of the same fact (e.g., March third, nineteen seventy-five, ``three three seventy-five,'' or ``zero three zero three one nine seven five'' for a DOB).

As illustrated in \autoref{fig:motivation}(a), current practice is largely \emph{data-first and human-in-the-loop}: teams wait for real phone transcripts to trickle in, then repeatedly hand-tune prompts on restricted Protected Health Information (PHI). Each new entity (ZIP, DOB, medications, etc.) restarts this slow cycle, and robustness is limited to the narrow linguistic patterns observed in the available calls. This workflow is further constrained by HIPAA, which makes real transcripts scarce, tightly access-controlled, and expensive to annotate \citep{pmc_2024_healthcare_voice_trust}. Existing NLP benchmarks such as CoNLL-2003 and clinical named entity recognition (NER) corpora \citep{tjong_kim_sang_2003_conll} focus on written text and do not capture the disfluencies and Linguistic Verbalization Patterns (LVPs)\footnote{A
\textit{linguistic verbalization pattern} is a recurring way in which a
structured value is realized in spoken language (e.g., different number
or date readings, fillers, hesitations, self-corrections).} of conversational healthcare speech.

We introduce \textbf{\textsc{LingVarBench}}\footnote{\textsc{LingVarBench} abbreviates \textit{Linguistic Variation Benchmark}; 
in this work, \textit{linguistic variation} and \textit{linguistic verbalization patterns} 
both refer to the same concept: different ways structured values are expressed in 
natural-language utterances.}, shown in \autoref{fig:motivation}(b) and detailed further in \autoref{fig:BenchFramework}, a semantic synthetic data generation and evaluation framework for healthcare voice AI. Our contributions are three-fold:
(1) we synthesize HIPAA-compliant patient--provider dialogues using large language models (LLMs) by combining entity descriptions with curated LVPs,e.g., digit-by-digit vs.\ grouped numbers, pauses, fillers, self-corrections;
(2) we use DSPy + SIMBA to automatically optimize extraction prompts on this synthetic data, avoiding manual prompt engineering on PHI (full framework in \autoref{fig:BenchFramework}); and (3) we provide a benchmark specification (entity schemas, LVP inventory, and generation/validation protocol) to measure robustness under controlled conversational variation.

On real-world production transcripts, prompts optimized exclusively on \textsc{LingVarBench} outperform zero-shot prompting and match—or even surpass—human-tuned prompts for structured entities such as ZIP, DOB, and name (F1 $\approx$ 94--95), despite the human prompts being crafted with access to real customer transcripts. For subjective questionnaire items, the optimized prompts remain comparable to human-tuned performance and substantially higher than zero-shot. These results show that a fully automated, synthetic, linguistically controlled pipeline can achieve near–human-level prompting without requiring access to PHI, providing a practical path to HIPAA-compliant healthcare voice AI. This approach is especially valuable at initial product launch, when real-world datasets are not yet available.

\section{Related Work}
Much recent work has focused on clean text and high-quality transcriptions for training their models \cite{arokodare2025clinical,liu2025using}.
These often fail to generalize effectively in noisy real-world scenarios. Moreover, these datasets frequently lack coverage of domain-specific or emerging entities, limiting their long-term applicability. For example, \citet{davidson-etal-2021-improved} reported low F1 scores for entities such as \textit{email address} (0.46) and \textit{named products} (0.65). \citet{9730848} and \citet{eger2020activation} show that even modern neural-based NER systems tend to memorize dataset-specific patterns rather than generalize. These models often break down across datasets due to dataset bias, annotation artifacts, and poor generalization to unseen entity types or domains—highlighting the shallow generalization of even well-curated datasets. 
Therefore, large datasets that accurately reflect real-world scenarios are crucial for effectively fine-tuning models and enhancing their performance~\cite{wang2024harmonic, bao-etal-2023-synthetic}.
% \citet{laskar-etal-2022-improving} and \citet{kaplan-2020-may} involved human annotation of product and organization entities from transcriptions to improve downstream extraction accuracy.
To overcome these shortcomings, synthetic data has been used to enhance model performance. Initially, generative adversarial networks (GANs)~\cite{NIPS2014_f033ed80} were popular for creating synthetic data closely resembling real samples. More recently, language models and other generative AI systems have been employed to create datasets for various training tasks. 
\citet{he-etal-2022-generate} introduced the Generate-Annotate-Learn (GAL) framework, which uses synthetic text data to improve classification performance. Similar approaches have been applied in other domains as well: for instance, synthetic speech has been used to train automatic speech recognition (ASR) models~\cite{9003990}, and synthetic visual data has supported semantic segmentation in autonomous driving~\cite{7780721}.
More recently, the emergence of LLMs has enabled the generation of high-quality synthetic labels, further enhancing model compression and distillation techniques. Fu et al. (2022) demonstrated this by using pseudo-labels from a fine-tuned teacher model, enabling a student model to achieve a 75 times speedup with only a 1\% drop in accuracy.

For model optimizations and fine-tuning, prompt-based learning has emerged as a powerful paradigm for adapting large models to downstream tasks \cite{brown2020language, wei2022chain}. However, prompt engineering for entity extraction remains a trial-and-error process, especially when dealing with unpredictable user input. Recent frameworks like DSPy \cite{khattab_2024_dspy} provide a principled way to optimize prompts using differentiable feedback, but their effectiveness is limited by the availability of diverse high-coverage datasets.

To train generalizable models with minimal human effort, we are proposing a synthetic dataset generation pipeline tailored for entity extraction from phone call transcriptions. We leverage LLMs to simulate realistic transcripts with controlled LVPs and use DSPy to optimize prompts that generalize across these variants. This enables rapid iteration and robust evaluation without relying on expensive human annotation.  In this work, we instantiate the pipeline using three commercially available LLMs: GPT 4, Gemini 2.0 Flash, and Gemini 2.5 Pro, allowing us to evaluate cross-model consistency and robustness.

\section{Methodology}

\subsection{Problem Definition}
Our problem formulation is grounded in the assumption that instruction-tuned language models implicitly model how humans realize structured values as spoken language, capturing diverse paraphrases, disfluency and surface forms for the same underlying value \cite{brown_2020_gpt3, ouyang2022training,zhang2025llm}. We operationalize this ``verbalization prior'' by conditioning the LLM on a value and a LVP, and then using it to sample candidate transcripts that we later validate for value consistency. Formally, given a target entity type \( e \in \mathcal{E} \), a structured field description \( d \in \mathcal{D} \), and a LVP \( v \in \mathcal{V} \), our goal is to generate a transcript \( u \in \mathcal{U} \) such that (i) the transcript \( u \) contains a valid instantiation of the entity \( e \), consistent with the field description \( d \); (ii) \(u\) reflects the specified linguistic variation \( v \); (iii) the distribution of \( u \) approximates the distribution of real-world phone call transcripts for the given entity and variation type. We model a novel function \( G \) as a composition of three components. 
The overall generative function is
\[
G(d, v) = \mathcal{C}\big( T( V(d), v ) \big),
\]
where \(G : (\mathcal{D}, \mathcal{V}) \rightarrow (\mathcal{U}, e)\),
\(V : \mathcal{D} \rightarrow \mathbb{V}\) is a \textit{Value Generator} that samples a plausible entity value from a field description,
\(T : \mathbb{V} \times \mathcal{V} \rightarrow \mathcal{U}\) is a \textit{Transcript Generator} that produces a natural-language utterance embedding the value with the desired variation,
and \(\mathcal{C} : \mathcal{U} \rightarrow (\mathcal{U}, e)\) is a \textit{Consistency Checker} that enforce value consistency and ensures the generated transcript is semantically correct and contains a recoverable entity \( e \).

% \[
% G(d, v) = \mathcal{C} \left( T \left( V(d),\ v \right) \right)
% \quad \text{where} \quad
% G: (\mathcal{D}, \mathcal{V}) \rightarrow (\mathcal{U}, e)
% \]

% \[
% G: (\mathcal{D}, \mathcal{V}) \rightarrow (\mathcal{U}, e)
% \]
% where \( G \) is implemented using a combination of large language model (LLM)-based value generation, variation-controlled prompt templating, and transcript validation to ensure semantic correctness and linguistic realism. 

\subsection{\textsc{LingVarBench} Overview}
% We modularize our pipeline into three main components: (1) Value Generation, (2) Transcript Generation, and (3) Transcript Validation. We use LLMs in each stage due to their flexibility, compatibility with optimization tools (e.g. DSPy), and strong performance in language understanding tasks.
We developed three key modules—Value Generator, Transcript Generator, and Consistency Checker, using LLMs. The data flow between these modules is illustrated in \autoref{fig:BenchFramework}. The process begins with the Value Generator, which takes a task description as input and produces plausible values aligned with the specified description. These generated values, along with predefined linguistic verbalization types, are then passed to the Transcript Generator, which constructs transcripts embedding the provided values. Finally, the Consistency Checker module verifies the plausibility and consistency between the generated transcripts and the corresponding values. Only those transcripts that pass this step are retained for use. We next highlight the key challenge and how our design enforces controllability and reliability.

Beyond simply prompting an LLM, the key technical challenge is ensuring that generated transcripts are both (i) linguistically diverse and (ii) semantically correct with respect to the intended structured value. In \textsc{LingVarBench}, controllability comes from explicitly composing each example from a field description and an LVP, while reliability comes from enforcing two constraints: distributional coverage (uniform coverage across value–variation pairs via recursive balancing) and semantic consistency (filtering out generations where the transcript does not unambiguously support the target value). Together, these constraints turn LLM generation into a repeatable benchmark construction procedure that supports fine-grained robustness evaluation and prompt optimization.
% \subsection{Components}
\subsubsection{Inputs}
We represent each field description using three components: the field name (e.g., ZIP code), the data type (e.g., integer), and a natural language description that specifies the format or constraints of the field (e.g., ``Zip code is a 5-digit number, with optional 4-digit add on code"). To better reflect real-world spoken language, we also define a set of LVPs that capture speech phenomena such as disfluencies, informal syntax, entity format variability, and self-corrections—features that are typically underrepresented in conventional benchmarks. 

In our experiments, we evaluated model performance on the following entities: ZIP code, DOB, name, pain rating, respiratory issues, and hearing issues. Together, these entities span a range of types, including integers, strings, dates, booleans, and multi-select enums. Although our experiments instantiate \textsc{LingVarBench} on these six entities, the framework is entity-agnostic: adding a new field requires only a schema-level description (name, type, and constraints) plus optional entity-specific LVP templates for common readout conventions. We view broader entity coverage (e.g., additional intake fields and open-vocabulary clinical concepts) as an important next step. We selected this set because it reflects the core fields used in our production intake flows: ZIP, DOB, and name are high-stakes authentication fields, while pain rating and respiratory/hearing issues are clinically relevant screening questions that stress both numerical and yes/no/multi-select reasoning under noisy speech. 

In this work, we focus on direct-answer turns, i.e., utterances that explicitly contain (or directly state) the target value in response to the system question. This isolates robustness to linguistic realization (LVPs) while keeping supervision well-defined. We do not yet model multi-turn resolution or dialogue acts such as refusals (``I’d rather not say''), topic shifts, clarification questions (``Which date do you mean?''), or pragmatic/implicit answers (``same as last time''), which are common in real phone calls and require additional dialogue-state or answerability modeling.

\subsubsection{Value Generation Module}
The Value Generation component uses an LLM prompted with the field description to produce sample values. The quantity of generated values is configurable, allowing for increased diversity in the output. An abstracted template prompt is shown in \autoref{fig:Prompt_ValueGeneration} (in \autoref{appendix_A}).
% Value Generation Prompt

\subsubsection{Transcript Generation Module}
The transcript generation modules generate all possible pairs of a) the values generated by the Value generation module and b) the LVPs. For each pair, the module prompts the LLM to generate a phone call transcription that will provide the specific value with the specific LVP. To prevent over-representation of certain value–variation pairs, the module employs a recursive generation strategy: underrepresented pairs are identified and re-prompted until a balanced distribution is achieved. This ensures uniform coverage across the dataset while preserving linguistic diversity. The output of this module is a list of transcripts for every generated value. An abstracted transcript-generation prompt appears in \autoref{fig:Prompt_TranscriptGeneration} (in \autoref{appendix_A}).

\paragraph{Linguistic Verbalization Patterns (LVPs)}
To simulate the variability of real-world speech, we define a set of LVPs applied during transcript generation. These patterns fall into two categories: (a) \textit{general variations}, which capture stylistic or pragmatic features such as disfluencies, hesitations, or confirmation-seeking, and (b) \textit{entity-specific variations}, which reflect how different structured fields (e.g., ZIP codes) are naturally expressed in spoken language. For example, the general variation \texttt{self-correction} is defined such that the model includes self-corrections in its response, e.g., \texttt{``it's one two... no wait, four five''}. Our LVPs model variation within direct answers; We isolate robustness to linguistic realization (LVPs) by mainly focusing on direct-answer turns, providing a controlled environment to measure information extraction accuracy specifically independently of multi-turn dialogue state modeling (refusals, topic shifts, multi-turn grounding), which allows for a more precise evaluation of the model's sensitivity to spoken-language variation.
% dialogue acts (refusals, topic shifts, multi-turn grounding) are left to future work (\S\ref{sec:limitations}).}

Entity-specific variations introduce structural diversity. For date-of-birth, for instance, we include types like \texttt{spoken\_date\_8\_digits}, which verbalizes the full date using individual spoken digits. An example would be: \texttt{"one two zero two one nine four seven"}. These curated types enable precise control over linguistic diversity and support fine-grained robustness evaluation for extraction models. A detailed description of all LVPs that we used is available in Tables~\ref{tab:variation-types-general}-\ref{tab:variation-types-respiratory-issues}, in \autoref{Appendix_D}. 

\paragraph{Design and extensibility of LVPs.}
We intentionally define LVPs as a lightweight, modular inventory (\autoref{Appendix_D}) rather than learning a latent variation model, because our goal is controllable stress-testing of extractors under specific spoken-language phenomena. The inventory is structured to be extensible: (i) \emph{general} LVPs capture broadly reusable dialogue surface phenomena (e.g., hesitation, self-correction, confirmation-seeking), while (ii) \emph{entity-specific} LVPs encode formatting conventions tied to an entity schema (e.g., digit grouping for DOB/ZIP, name variants). To extend \textsc{LingVarBench} to a new domain or language, one typically keeps the general LVPs and adds/edits a small set of entity-specific templates reflecting local conventions; the rest of the pipeline remains unchanged. Newly added LVPs can be sanity-checked automatically by generating samples and retaining only those where the intended value is recoverable under the same consistency validation used in the pipeline.

\subsubsection{Transcript Consistency Checker Module}
Recent work shows that LLMs can behave unpredictably under knowledge conflicts, sometimes readily incorporating external evidence and sometimes stubbornly adhering to their internal parametric memory \cite{xie2023adaptive, mohammadi2025llms}. Motivated by these findings, we incorporate a \textit{Transcript Consistency Checker} module within the pipeline. This module reuses the same LLM as a verifier to determine whether a generated transcript correctly contains the intended value. To mitigate the impact of false positives on the data distribution, the system recursively invokes 
the Transcript Generation Module until the target number of valid samples is achieved. In our current setup, the checker filters out non-answer turns (e.g., refusals or off-topic responses), reflecting our direct-answer scope. Although false negatives pose a risk by introducing labeling noise, this can be mitigated through more stringent validation criteria (e.g., requiring exact value recovery or higher confidence).

\subsection{Prompt Optimization with DSPy}
For every entity, we used the \textsc{LingVarBench} Synthetic Data Generation framework to generate about a thousand labels of ground truth data. This was then split into training and test dataset. We used the training data to optimize the prompts for entity extraction using DSPy \cite{khattab_2024_dspy} and SIMBA optimization engine. The optimized prompt was then tested with the test split of the dataset. 
We then used our proprietary real phone call dataset from calls to patients to verify that the optimized prompt works in the real world. The base instruction used for DSPy optimization is shown in \autoref{fig:Prompt_DSP} (\autoref{appendix_A}).

\section{Experimental Setup and Results}
\subsection{Proprietary Dataset}
We deploy voice AI agents to automate patient-facing phone calls in healthcare settings. Every call is reviewed by a human-in-the-loop before submitting to the customer.
Our customers also audit the data for correctness providing us with a second layer of validation. For each question asked by the voice agent, the dataset includes the corresponding patient utterance and the extracted entity response. 
The two layer validation process leads to a high-quality ground-truth dataset for entity extraction from real-world phone interactions. 
Due to the sensitive nature of healthcare data, this dataset cannot be open-sourced or publicly released. However, we used it internally to validate the performance of our optimized prompts. \autoref{fig:exampledataset}, in Appendix A, presents fabricated examples to illustrate the structure and content of the dataset.

\paragraph{Data availability and reproducibility.}
Due to privacy, consent, and organizational constraints, we do not release the proprietary real-call dataset used for external validation, the generated synthetic benchmark dataset, or our internal codebase. We therefore provide a detailed benchmark specification (entity schemas and LVP inventory in \autoref{Appendix_D}), generation/validation protocol, prompt templates in \autoref{appendix_A}, and complete experimental settings to support independent reimplementation and analysis.

\subsection{Metrics}
To evaluate the effectiveness of our approach, we conducted experiments to measure (i) the entity extraction accuracy achieved using prompts optimized with \textit{LingVarBench}-generated data; (ii) the similarity between real transcripts and \textsc{LingVarBench}-generated synthetic transcripts.

\paragraph{Entity extraction accuracy}
We used \textsc{LingVarBench} to optimize entity extraction prompts for three general entities, zip code, name, DOB, and three specific entities, pain rating, respiratory issues, and hearing issues, for which human-labeled data are available.  We compared the accuracy of prompts optimized using \textsc{LingVarBench} against two baselines: (a) a zero-shot human-written prompt, and (b) a human-written prompt optimized through evaluation and tuning on historical data.
We compute F1 scores using binary correctness at the sample level.

\paragraph{Similarity scores}

\begin{table*}[t]
    \centering
    \footnotesize
    \renewcommand{\arraystretch}{0.9}
    \begin{tabularx}{\textwidth}{l l *{12}{>{\centering\arraybackslash}X}}
        \toprule
        \multirow{2}{*}{\makecell{\textbf{Prompt}\\\textbf{Type}}} & \multirow{2}{*}{\textbf{Model}} 
            & \multicolumn{2}{c}{\textbf{ZIP(\%)}} 
            & \multicolumn{2}{c}{\textbf{Name(\%)}} 
            & \multicolumn{2}{c}{\textbf{DOB(\%)}}
            & \multicolumn{2}{c}{\textbf{Pain Rt.(\%)}}
            & \multicolumn{2}{c}{\textbf{Resp. Iss.(\%)}}
            & \multicolumn{2}{c}{\textbf{Hearing Iss.(\%)}}\\
        \cmidrule(lr){3-4} \cmidrule(lr){5-6} \cmidrule(lr){7-8} \cmidrule(lr){9-10} \cmidrule(lr){11-12}\cmidrule(lr){13-14}
            & & \textbf{Acc} & \textbf{F1} 
              & \textbf{Acc} & \textbf{F1} 
              & \textbf{Acc} & \textbf{F1} 
              & \textbf{Acc} & \textbf{F1}
              & \textbf{Acc} & \textbf{F1}
              & \textbf{Acc} & \textbf{F1}\\
        \midrule
        \multirow{3}{*}{0-shot}
            & GPT 4             & 88.62 & 93.93 & 78.19 & 87.77 & 74.52 & 85.42 & 80.18 & 81.41 & 96.18 & 95.04  & 87.43 & 87.88 \\
            & Gem 2.5     & 89.37 & 94.34 & 79.15 & 88.38 & 77.04 & 87.01 & 82.28 & 85.10 & 96.84 & 95.77 & 82.94 & 83.70\\
            & Gem 2.0  & 88.50 & 93.85 & 47.87 & 64.75 & 72.50 & 84.04 & 80.97 & 82.51 & 96.18 & 94.92 & 84.18 & 84.84\\
        \midrule
        \multirow{3}{*}{Human}
            & GPT 4        & 88.87 & 94.06 & 79.16 & 88.38 & 80.28 & 89.05 & \textbf{86.22} & \textbf{87.33} & 81.18 & 63.50 & \textbf{89.79} & \textbf{90.09 } \\
            & Gem 2.5    & 89.75 & 94.54 & 79.56 & 88.59 & 81.84 & 90.00 & 82.15 & 85.36 & \textbf{97.50} & \textbf{96.65}  & 82.94 & 83.61\\
            & Gem 2.0  & 89.76 & 94.58 & 83.38 & 90.97 & \textbf{82.62} & 90.47 & 81.10 & 82.7 & 97.11 & 96.04 & 84.06 & 84.65\\
        \midrule
        \multirow{3}{*}{\makecell[l]{Optimized\\ (Ours)}}
            & GPT 4             & 89.79 & \textbf{94.66} & \textbf{90.18} & \textbf{94.84} & 78.59 & \textbf{94.84} & 80.05 & 81.53 & 96.71 & 95.67 & 88.44 & 88.74   \\
            & Gem 2.5    & \textbf{95.07} & 94.61 & 85.20 & 94.15 & 80.80 & 89.38 & 82.94 & 85.93 & 96.58 & 95.56 & 88.78 & 89.15\\
            & Gem 2.0  & 89.14 & 89.30 & \textbf{90.18} & \textbf{94.84} & 80.66 & 89.30 & 85.43 & 86.8 & 96.84 & 95.86 & 88.10 & 88.45\\
        \bottomrule
    \end{tabularx}
    \caption{Accuracy and F1 on real-world production data using zero-shot, human-written (tuned on real data for Gemini 2.0), and optimized prompts. Our prompts are optimized using only synthetic data generated by \textsc{LingVarBench}, yet outperform zero-shot and match or exceed human-tuned prompts for ZIP and Name, and improve over 0-shot for DOB.}
   \label{tab:promptmodelaccuracy}
\end{table*}

We used text embedding models in conjunction with cosine similarity to assess the similarity between \textsc{LingVarBench}-generated transcripts and real phone call transcripts. 
\begin{equation*}
\text{Similarity}(T_\text{real}, T_\text{synthetic}) = \frac{\vec{E}_{\text{real}} \cdot \vec{E}_{\text{synthetic}}}{\|\vec{E}_{\text{real}}\| \, \|\vec{E}_{\text{synthetic}}\|}
\label{eq:cosine_similarity}
\end{equation*}
\noindent where:
    $T_\text{real}$ is the real phone call transcript annotated with ground truth entities,
    $T_\text{synthetic}$ is the corresponding synthetic transcript generated by \textsc{LingVarBench} using the same entity values and LVP, 
    $\vec{E}_{\text{x}}$ is the embedding vector of the input text.

\begin{table}[t]
    \centering
    \scriptsize
    \setlength{\tabcolsep}{2pt}       % less horizontal padding
    \renewcommand{\arraystretch}{0.95} % slightly tighter rows
    \begin{tabularx}{\columnwidth}{ll*{6}{>{\centering\arraybackslash}X}}
        \toprule
        \textbf{Model} & \textbf{Var.} & \textbf{ZIP} & \textbf{Name} & \textbf{DOB} & \textbf{Pain} & \textbf{Resp.} & \textbf{Hear.} \\
        \midrule
        \multirow{4}{*}{text-embedding-3-large}
          & Match    & \textbf{0.81} & \textbf{0.81} & \textbf{0.62} & \textbf{0.66} & \textbf{0.57} & \textbf{0.69} \\
          & Super    & 0.78 & 0.77 & 0.58 & 0.65 & 0.55 & 0.58 \\
          & Sub      & 0.72 & 0.65 & 0.62 & 0.64 & 0.53 & 0.59 \\
          & Null     & 0.67 & 0.55 & 0.55 & 0.42 & 0.43 & 0.39 \\
        \midrule
        \multirow{4}{*}{gemini-embedding-001}
          & Match    & \textbf{0.91} & \textbf{0.92} & \textbf{0.83} & \textbf{0.75} & \textbf{0.65} & \textbf{0.78} \\
          & Super    & 0.90 & 0.91 & 0.82 & 0.72 & 0.62 & 0.70 \\
          & Sub      & 0.87 & 0.83 & \textbf{0.83} & 0.71 & 0.62 & 0.71 \\
          & Null     & 0.85 & 0.77 & 0.80 & 0.70 & 0.58 & 0.67 \\
        \bottomrule
    \end{tabularx}
\caption{Cosine similarity between real and synthetic transcripts. Let $V_r$ = variation types in real transcripts, $V_s$ = variation types in synthetic transcripts. Match refers to the similarity scores when $V_s = V_r$; Super refers to the similarity scores when  $V_s \supseteq V_r$ ; Sub refers to the similarity scores when $V_s \subseteq V_r$; and Null refers to the case when: $V_s \cap V_r = \emptyset$. Full results with standard deviations are in \autoref{tab:similarity_scores_models_appndx} in Appendix~C.}

% \caption{Table 2: Cosine similarity between real and synthetic transcripts across variation types (higher is better). "Match" = synthetic transcripts use identical variation types as real transcripts; "Super" = synthetic variations are a superset of real variations; "Sub" = synthetic variations are a subset of real variations; "Null" = no overlap between synthetic and real variation types. Full results with standard deviations are in Table 5 in Appendix C.Retry Full results with standard deviations are, \autoref{tab:similarity_scores_models_appndx} in Appendix~C.}
    \label{tab:similarity_scores_models}
\end{table}

\subsection{Implementation Details}
We implemented the pipeline in Python using DSPy \cite{khattab_2024_dspy}. 
We instantiate the pipeline using three commercially available LLMs: GPT 4 (via Azure OpenAI), Gemini 2.0 Flash, and Gemini 2.5 Pro (via Google Vertex AI). All models are used across the generation, validation, and extraction stages. Prompt formats were standardized across models, and decoding parameters (temperature = 0, top-p = 1.0) were kept consistent. This setup allows us to evaluate the framework's robustness across models. We chose GPT-4 and Gemini 2.5 Pro because they are the state of the art cloud-hosted models for which we have appropriate  Business Associate Agreements (BAAs). We tested with Gemini 2.0 flash because we leverage faster LLMs in production to deliver low-latency experience to patients.
 For semantic similarity evaluations, we used the \texttt{text-embedding-3-large} (OpenAI) and \texttt{gemini-embedding-001} (Google) models.

\subsection{Results and Discussion}
We evaluate \textsc{LingVarBench} on six entities for which our proprietary dataset provides high-quality, human-validated ground truth. We group these into \emph{general entities} (ZIP code, DOB, name) and \emph{task-specific entities} (pain rating, respiratory issues, hearing issues). Together, they span integers, strings, dates, booleans, and multi-select enums, and cover both common patient-identifying information and clinically relevant screening questions. We focus on this set because it is the subset for which our production evaluation data provides consistent, human-validated ground truth. Our framework is designed to be entity-agnostic. Although we demonstrate its efficacy on a core set of six critical healthcare and authentication fields, the architecture allows for seamless extension to new domains and tasks via schema-level descriptions for future work. Summary statistics of the synthetic transcripts and DSPy training/validation/test splits are reported in \autoref{Appendix_B} (Tables~\ref{tab:dataset_summary} and~\ref{tab:performance}).

\autoref{tab:promptmodelaccuracy} reports the performance of zero-shot, human-written, and \textsc{LingVarBench}-optimized prompts on real production calls. For general entities, \textsc{LingVarBench}-optimized prompts trained \emph{only} on synthetic data achieve approximately 90\% accuracy for ZIP and name, matching or exceeding human-tuned prompts. DOB accuracy is somewhat lower, largely due to synthetic samples that use less common date formats (e.g., \texttt{dd-mm-yyyy}) in U.S. speech, highlighting the importance of domain-aware formatting priors when generating synthetic transcripts. 

Among task-specific entities, pain rating is more challenging because patients often respond with ambiguous phrases such as ``none'', ``high'', or ``low'', which reduces accuracy despite strong performance on canonical numerical answers. Respiratory issues are asked via a yes/no question (``Do you have any respiratory issues?''), yielding generally high accuracy and F1. Hearing issues are selected from a small, fixed set of options; because the response space is highly constrained, variation is limited and all prompt types perform similarly.

In real-world conversations, entities may not appear immediately or directly in response to a question, introducing labeling noise because human annotators do not mark the exact utterance span; as a result, 100\% accuracy is effectively unattainable. Moreover, the human-crafted prompt was optimized specifically for Gemini~2.0~Flash, so it is expected to outperform other models when re-used without further tuning.

\autoref{tab:similarity_scores_models} presents cosine similarity between synthetic and real transcripts with matched entity values. We observe that similarity increases as the alignment of LVPs between real and synthetic transcripts improves, and this trend holds across both embedding models. This supports that our synthetic generation pipeline produces transcripts that closely resemble real-world conversational data, even though both sides share the same underlying entity values. Additional similarity results with standard deviations are included in the Appendix. We hypothesize that incorporating an even broader range of LVPs could further sharpen the separation between variation-matched and variation-mismatched conditions.

\section{Conclusion}

This work presents \textsc{LingVarBench}, a synthetic data generation pipeline leveraging large language models to produce NER datasets tailored for phone call transcripts, guided by human-defined entity specifications and linguistic verbalization patterns. We demonstrate that entity extraction models trained on \textsc{LingVarBench}-generated data for various entities achieve performance comparable to that obtained via human-crafted prompt tuning on large-scale real-world call datasets.

\section{Limitations}
\label{sec:limitations}

The framework presently models only direct answers that explicitly respond to the question; it does not yet capture indirect, ambiguous, or off-topic utterances (e.g., refusals, topic shifts, or pragmatic answers). Extending the framework to simulate such complex dialogue phenomena remains an important direction for future work.

Our LVP inventory is curated for controllable variation but is not exhaustive; expanding to new domains or languages may require new templates. We also evaluate six structured entities with reliable ground truth, and leave broader coverage (e.g., additional intake fields or open-vocabulary concepts) to future work.

We note that the human-optimized prompt underperformed relative to the zero-shot GPT-4 prompt when extracting respiratory-issue entities, because the human prompt was optimized for the Gemini 2.0 Flash model.

\section*{Ethical Considerations}
This research centers on using large language models to generate synthetic data for training NER systems tailored to phone-call transcripts. We do not release any real customer transcripts, and we also do not release the generated synthetic benchmark dataset due to organizational data-sharing constraints. To support transparency, we document the full benchmark specification (entity schemas and LVP inventory) and the generation/validation protocol.

\section*{Acknowledgments}
We thank Dr. Manas Gaur (Prof. at University of Maryland, Baltimore County) for providing valuable feedback on the manuscript. We also thank Amanda Griffin for her assistance with preparing the figures. Finally, we thank Infinitus System, Inc. for providing access to the real-world production transcripts used for evaluation, as well as for the financial support and computational resources that enabled this research.

\bibliography{custom}

\clearpage
\appendix
\section{Prompt Templates} % Appendix A
\label{appendix_A}

%Abstracted Prompt for Value Generation
\begin{figure}[!h]
    \centering
    {\scriptsize
\begin{tcolorbox}[
  colback=gray!2, colframe=black!30, title=\scriptsize \textbf{Abstracted Prompt for Value Generation},
  fonttitle=\scriptsize,
  coltitle=black,
  boxrule=0.2mm,
  arc=1mm,
  left=1mm,
  right=1mm,
  top=1mm,
  bottom=1mm,
  enhanced,
  breakable
]
% --- top content ---
\textbf{Input Fields:}
\begin{itemize}[itemsep=0pt, parsep=0pt]
  \item \textbf{Field Name}: \texttt{\{field\_name\}}
  \item \textbf{Field Description}: \texttt{\{field\_description\}}
  \item \textbf{Question}: \texttt{\{question\}}
  \item \textbf{Expected Output Type}: \texttt{\{output\_type\}}
  \item \textbf{Number of Values}: \texttt{\{num\_values\}}
\end{itemize}

% --- bottom block starts here ---
\tcblower
\textbf{Output Format (JSON):}
\begin{verbatim}
{ "values": [ "value_1", "value_2","...", value_N" ] }
\end{verbatim}
\end{tcolorbox}
}
    \caption{Abstracted Prompt for Value Generation}
    \label{fig:Prompt_ValueGeneration}
\end{figure}

%Transcript generation prompt
\begin{figure}[!h]
    \centering
    {\scriptsize
\begin{tcolorbox}[colback=gray!2, colframe=black!30, title=\textbf{Abstracted Prompt for Transcript Generation},   fonttitle=\scriptsize,
  coltitle=black,
  boxrule=0.2mm,
  arc=1mm,
  left=1mm,
  right=1mm,
  top=1mm,
  bottom=1mm,
  enhanced,
 breakable]

\textbf{Input Fields:}
\begin{itemize}[itemsep=0pt, parsep=0pt]
  \item  \textbf{Question}: \texttt{\{question\}}
  \item  \textbf{Output Type}: \texttt{\{output\_type\}}
  \item  \textbf{Target Value}: \texttt{\{ground\_truth\}}
  \item  \textbf{ Existing Transcripts}: \texttt{\{existing\_transcript\_list\}}
  \item  \textbf{Variation Types}: \texttt{\{variation\_types\_str\}}
  \item  \textbf{Variation Instructions}: \texttt{\{variation\_descriptions\_str\}}
\end{itemize}

\tcblower

\textbf{Task:} Generate additional \textit{natural spoken transcripts} that verbalize the target value without altering its meaning.

\textbf{Key Constraints:}
\begin{itemize}[itemsep=0pt, parsep=0pt]
  \item  Always express the target value (\texttt{\{ground\_truth\}}) in natural spoken form
  % \item  Avoid literal strings, quotes, or random substitutions
  % \item  Do not duplicate any \texttt{existing\_transcript\_list}
  \item  For dates: include both spoken and digit-only formats (e.g., \texttt{"January fifth, 1989"} and \texttt{"1589"})
  \item  For names: add a realistic last name to the first name
\end{itemize}
\textbf{Variation Type Assignment:}
\begin{itemize}[itemsep=0pt, parsep=0pt]
  \item  Assign \textbf{one or more} variation types per transcript from \texttt{\{variation\_types\_str\}}
  \item  Ensure even distribution across all variation types
  \item  Use \texttt{"not\_listed"} if the transcript doesn't match any type
\end{itemize}
\textbf{Diversity Rule:} Be creative in \textit{how} the value is spoken, but do not change \textit{what} the value is.
\vspace{0.5em}
\textbf{Output Format (JSON):}
\begin{verbatim}
{  "transcripts":  
    [ { "transcript": "spoken response",
      "variation_types": ["type1","type2"]} ]
 }
\end{verbatim}
\textbf{Output Constraints:}
\begin{itemize}[itemsep=0pt, parsep=0pt]
  \item  Return only valid JSON — no markdown, no extra text
  \item  Each transcript must clearly verbalize the value
  \item  All responses must match the specified output type
\end{itemize}

\end{tcolorbox}}
    \caption{Abstracted Prompt for Transcript Generation}
    \label{fig:Prompt_TranscriptGeneration}
\end{figure}

% Validation Prompt
\begin{figure}[!t]
    \centering
    {\scriptsize
\begin{tcolorbox}[colback=gray!2, colframe=black!30, title=\textbf{Abstracted Prompt for Validation},   fonttitle=\scriptsize,
  coltitle=black,
  boxrule=0.2mm,
  arc=1mm,
  left=1mm,
  right=1mm,
  top=1mm,
  bottom=1mm,
  enhanced,
  breakable
]
\textbf{Input Fields:}
\begin{itemize}[itemsep=0pt, parsep=0pt]
  \item \textbf{Transcript}: \texttt{\{transcript\}}
  \item \textbf{Ground Truth}: \texttt{\{ground\_truth\}}
  \item \textbf{Action Name}: \texttt{\{action\_name\}}
\end{itemize}

\tcblower

\textbf{Task:} Determine if the ground truth value can be extracted from the transcript.\\
\textbf{Rules:}
\begin{itemize}[itemsep=0pt, parsep=0pt]
  \item If the transcript contains the value (even with corrections) → \texttt{true}
  \item If the transcript is vague or doesn't contain the value → \texttt{false}
  \item If the value is mentioned at any point → \texttt{true}
\end{itemize}
\textbf{Examples:}
\begin{itemize}[itemsep=0pt, parsep=0pt]
  \item \texttt{"My zip is one two three four five"}, \texttt{12345} → \texttt{true}
  \item \texttt{"I don't know"}, \texttt{12345} → \texttt{false}
  \item \texttt{"seven oh ... no, nine oh two one oh"}, \texttt{90210} → \texttt{true}
\end{itemize}
\textbf{Date Format Considerations:}
\begin{itemize}[itemsep=0pt, parsep=0pt]
  \item Accept continuous digit formats: \\ e.g., \texttt{01-15-2024} → \texttt{01152024}, \texttt{11524}, etc.
  \item Spoken digit sequences: \\ e.g., \texttt{"zero one one five two zero two four"} are valid
\end{itemize}
\textbf{Output:} \texttt{true} or \texttt{false}
\end{tcolorbox}
}
    \caption{Abstracted Prompt for Validation}
    \label{fig:Prompt_Validation}
\end{figure}

% DSP Prompt
\begin{figure}[!t]
    \centering
    {\scriptsize
\begin{tcolorbox}[colback=gray!2, colframe=black!30, title=\textbf{Base Extraction Instructions used in DSPy optimization (ZIP Code)}, fonttitle=\scriptsize,
  coltitle=black,
  boxrule=0.2mm,
  arc=1mm,
  left=1mm,
  right=1mm,
  top=1mm,
  bottom=1mm,
  enhanced,
  breakable
]

\textbf{General Extraction Instructions:}

\begin{itemize}[itemsep=0pt, parsep=0pt]
  \item Extract the value from the transcript using the given question, field type, and field description.
  \item \textbf{Question}: \texttt{\{question\}}
  \item \textbf{Field Type}: \texttt{\{output\_type\}}
  \item \textbf{ Field Description}: \texttt{\{action\_description\}}
\end{itemize}
\textbf{Output Format:}
\begin{itemize}[itemsep=0pt, parsep=0pt]
  \item Return only the extracted value
  \item Do not include any symbols, labels, or extra text
\end{itemize}
\textbf{Example:}
\begin{verbatim}
Input: transcript: "...", Output: predicted: "..."
\end{verbatim}
\vspace{0.5em}
\textbf{ZIP Code Specific Guidelines:}
\begin{itemize}[itemsep=0pt, parsep=0pt]
  \item Extract exactly \textbf{5 numeric digits}
  \item Do not interpret ZIP codes as dates
  \item Return the raw 5-digit number only
  \item Example: \texttt{"one two three four five"} → \texttt{"12345"}
\end{itemize}

\end{tcolorbox}
}
    \caption{Base Extraction Instructions used in DSPy optimization (ZIP Code)}
    \label{fig:Prompt_DSP}
\end{figure}

\begin{figure}[th]
\small
\begin{quote}
\textbf{Agent:} Could you tell me your name please?\\
\textbf{Patient:} It is John.\\
\textit{→ Entity extracted:} \textbf{Name} \quad \textit{Entity value:} \textbf{John}

\vspace{0.6em}
\textbf{Agent:} What is your zip code?\\
\textbf{Patient:} It is nine double one oh one.\\
\textit{→ Entity extracted:} \textbf{ZIP Code} \quad \textit{Entity value:} \textbf{91101}
\textbf{Agent:} Do you have any respiratory issues?\\
\textbf{Patient:} Yes, I have asthma.\\
\textbf{Agent:} What about any of these hearing issues? Deafness, hard of hearing, or do you use hearing aids?\\
\textbf{Patient:} Hearing aid.\\
\end{quote}
\caption{Fabricated examples reflecting the structure of real patient–agent interactions used in evaluation. Real transcripts cannot be shown due to HIPAA constraints.}
\label{fig:exampledataset}
\end{figure}

\onecolumn

\section{Dataset and Implementation Details}  % Appendix B
\label{Appendix_B}
\begin{table*}[!h]
    \centering
    \footnotesize
    \renewcommand{\arraystretch}{1.1}
    \begin{tabularx}{\textwidth}{l l l l l l l l}
        \toprule
        \textbf{Entity} & \textbf{Model} & \textbf{\# Samples} & \textbf{Split (Train/Valid/Test)} & \textbf{\# Tags} & \textbf{\# Tag Occ.} & \textbf{\# Values} & \textbf{Avg Len (±std)} \\
        \midrule
        \multirow{4}{*}{ZIP code}
            & GPT 4  & 5635 & 3944 / 845 / 846     & 33 & 12449 & 10 & 41.7 ± 10.0 \\
            & Gem 2.5    & 1332   & 932 / 199 / 201                  & 33 & 2539    & 5 & 41.3 ± 20.0 \\
            & Gem 2.0  & 6467 & 4526 / 970 / 971     & 31 & 12101 & 12 & 41.1 ± 11.2 \\
            % & \textit{Entity total/avg} & \textit{--} & \textit{-- / -- / --} & \textit{--} & \textit{--} & \textit{--} & \textit{--} \\
        \midrule
        \multirow{4}{*}{Name}
            & GPT 4 & 2550 & 1785 / 382 / 383     & 46 & 5466 & 12 & 26.0 ± 8.5 \\
            & Gem 2.5 & 2164   & 1514 / 324 / 326 & 51  & 3468    & 5 & 29.7 ± 11.1 \\
            & Gem 2.0 & 6631 & 4641 / 994 / 996     & 46 & 12397 & 12 & 31.3 ± 13.0 \\
            % & \textit{Entity total/avg} & -- & -- & -- & -- & -- & -- \\
        \midrule
        \multirow{4}{*}{DOB}
            & GPT 4 & 1055   & 739 / 158 / 158                   & 36 & 2314    & 7 & 54.63 ± 10.74 \\
            & Gem 2.5  & 821   & 574 / 123 / 124                   & 34 & 1436    & 3 & 48.66 ± 19.31 \\
                        & Gem 2.0  & 682  & 477 / 102 / 103      & 33 & 682   & 5  & 63.3 ± 34.0 \\
            \midrule
            \multirow{4}{*}{Pain Rt.}
            & GPT 4 & 296 & 207 / 44 / 45 & 19 & 490 & 11 & 39.7 ± 16.8 \\
            & Gem 2.5 & 885 & 619 / 132 / 134 & 19 & 1055 & 11 & 38.0 ± 16.6 \\
            & Gem 2.0 & 823 & 576 / 123 / 124 & 19 & 1082 & 11 & 39.4 ± 17.8 \\
            \midrule
            \multirow{4}{*}{Resp. Iss.}
            & GPT 4 & 2114 & 1479 / 317 / 318 & 21 & 4056 & 2 & 56.4
  $\pm$ 20.6 \\
            & Gem 2.5 & 886 & 620 / 132 / 134 & 21 & 1111 & 2 & 40.9
  $\pm$ 20.5 \\
            & Gem 2.0 & 704 & 492 / 105 / 107 & 22 & 886 & 2 & 61.9 $\pm$
   38.0 \\  
            \midrule
            \multirow{4}{*}{Hearing Iss.}
            & GPT 4 & 795 & 556 / 119 / 120 & 37 & 861 & 7 & 45.5 ± 19.2 \\
            & Gem 2.5 & 992 & 694 / 148 / 150 & 37 & 1733 & 7 & 57.7 ± 25.5 \\
            & Gem 2.0 & 4530 & 3171 / 679 / 680 & 37 & 5846 & 7 & 43.6 ± 19.5\\
         \bottomrule
    \end{tabularx}
    \caption{Dataset statistics by entity and model. Totals or averages are reported per entity and across all entities for applicable columns The total number of valid samples varies because, although we request five samples per prompt per iteration, this is not strictly enforced. Hence, the LLM may produce fewer than five samples. The proportion of invalid samples was below 1\%. Note that ``\# Tag Occ.'' is the total count of variation tags assigned across all transcripts. Since transcripts can have multiple tags, this exceeds the number of samples. }
    \label{tab:dataset_summary}
\end{table*}

\begin{table}[!h]
    \small  % Reduce font size to help it fit
    \centering
    \renewcommand{\arraystretch}{0.95}
    \begin{tabularx}{0.5\columnwidth}{l l l l l}
        \toprule
        \textbf{Entity} & \textbf{Model} & \textbf{Valid Acc.(\%)} & \textbf{Test Acc.(\%)}\\
        \midrule
        \multirow{3}{*}{ZIP code}
            & GPT 4  & 96.92  & 96.93 \\
            & Gem 2.5 & 80.90  & 78.11 \\
            & Gem 2.0 &  89.59 & 89.19 \\
        \midrule
        \multirow{3}{*}{Name}
            & GPT 4  & 96.34 &  97.34 \\
            & Gem 2.5 & 72.22 &  70.25 \\
            & Gem 2.0 & 79
            .58& 79.51 \\
        \midrule
        \multirow{3}{*}{DOB}
            & GPT 4  & 98.10 & 98.11  \\
            & Gem 2.5 & 100 &  97.58 \\
            & Gem 2.0          & 94.12  & 99.03 \\
        \midrule
        \multirow{3}{*}{Pain Rt.}
            & GPT 4  & 100 & 100  \\
            & Gem 2.5 & 100 & 100 \\
            & Gem 2.0  & 99.19 & 100 \\
        \midrule
        \multirow{3}{*}{Resp. Iss.}
            & GPT 4  & 100 & 100 \\
            & Gem 2.5 & 100 & 96.58 \\
            & Gem 2.0 & 97.29 & 100 \\
        \midrule
        \multirow{3}{*}{Hearing Iss.}
            & GPT 4  & 68.91 & 79.17\\
            & Gem 2.5 & 84.46 & 84.91\\
            & Gem 2.0  & 78.64 & 80.44 \\
        \bottomrule
    \end{tabularx}
    \caption{Validation and test accuracy on synthetic data during DSPy-based prompt optimization. All results are based on \textsc{LingVarBench}-generated transcripts for each entity type.}
    \label{tab:performance}
\end{table}

\newpage

\section{Additional Results}  % Appendix C
\label{Appendix_C}
\begin{table}[!h]
    \centering
    \footnotesize
    \begin{subtable}{.8\columnwidth}
        \centering
        \begin{tabularx}{\columnwidth}{l *{6}{>{\centering\arraybackslash}X}}
            \toprule
            \textbf{Variation Type} & \textbf{ZIP code} & \textbf{Name} & \textbf{DOB} & \textbf{Pain Rt.} & \textbf{Resp. Iss.} & \textbf{Hearing Iss.}\\
            \midrule
            Match & \textbf{0.81}$\pm$0.13 & \textbf{0.81}$\pm$0.15 & \textbf{0.62}$\pm$0.14 
            & \textbf{0.66}$\pm$0.10 & \textbf{0.57}$\pm$0.02 & \textbf{0.69}$\pm$0.2  \\
            Superset variation         & 0.78$\pm$0.13 & 0.77$\pm$0.15 & 0.58$\pm$0.17 
                        & 0.65$\pm$0.10  & 0.55$\pm$0.02 & 0.58$\pm$0.18 \\
            Subset variation           & 0.72$\pm$0.14 & 0.65$\pm$0.17 & 0.62$\pm$0.14 
                        & 0.64$\pm$0.15  & 0.53$\pm$0.03 & 0.59$\pm$0.20  \\
            Null overlap               & 0.67$\pm$0.14 & 0.55$\pm$0.17 & 0.55$\pm$0.17 
                        & 0.42$\pm$0.15  & 0.43$\pm$0.03  & 0.39$\pm$0.21\\
            \bottomrule
        \end{tabularx}
        \subcaption{\texttt{text-embedding-3-large} (OpenAI)}
    \end{subtable}

    \vspace{1em}

    \begin{subtable}{.8\columnwidth}
        \centering
        \begin{tabularx}{\columnwidth}{l *{6}{>{\centering\arraybackslash}X}}
            \toprule
            \textbf{Variation Type} & \textbf{ZIP code} & \textbf{Name} & \textbf{DOB} & \textbf{Pain Rt.} & \textbf{Resp. Iss.} & \textbf{Hearing Iss.}\\
            \midrule
            % source https://docs.google.com/spreadsheets/d/1LX48NvmYF8tPqZGIrlj9u171QlLrDFN5oqxw7amDW1E/edit?gid=936212672#gid=936212672
            Match & \textbf{0.91}$\pm$0.07 & \textbf{0.92}$\pm$0.08 & \textbf{0.83}$\pm$0.08
                        & \textbf{0.75}$\pm$0.05 & \textbf{0.65}$\pm$0.02  & \textbf{0.78}$\pm$0.11 \\
            Superset variation         & 0.90$\pm$0.07 & 0.91$\pm$0.1 & 0.82$\pm$0.09 
                        & 0.72$\pm$0.05 & 0.62$\pm$0.03 & 0.70$\pm$0.11 \\
            Subset variation           & 0.87$\pm$0.07 & 0.83$\pm$0.08 & \textbf{0.83}$\pm$0.09 
                        & 0.71$\pm$0.05 & 0.62$\pm$0.02& 0.71$\pm$0.11 \\
            Null overlap               & 0.85$\pm$0.07 & 0.77$\pm$0.1 & 0.80$\pm$0.09 
                        & 0.70$\pm$0.06 & 0.58$\pm$0.02 & 0.67$\pm$0.11 \\
            \bottomrule
        \end{tabularx}
        \subcaption{\texttt{gemini-embedding-001} (Google)}
    \end{subtable}
    \caption{Similarity scores across different embedding models. 
    ``Match'' indicates identical variation types in both generated and reference transcripts. 
    ``Superset variation'' refers to generated transcripts containing all variation types from the reference plus additional ones. 
    ``Subset variation'' uses a non-empty subset of the reference's variation types. 
    ``Null overlap'' indicates no shared variation types.}
    \label{tab:similarity_scores_models_appndx}
\end{table}

\section{Linguistic Verbalization Patterns}  % Appendix D
\label{Appendix_D}

\paragraph{Extending the LVP inventory (domains, languages, new phenomena).}
Each LVP is specified as a short \emph{instruction template} plus an \emph{example} ((Tables~\ref{tab:variation-types-general}--\ref{tab:variation-types-respiratory-issues})
). Extending the inventory is straightforward: (1) reuse general LVPs (disfluency, self-correction, confirmation cues) across entities and domains; (2) add entity-specific LVPs by enumerating plausible surface realizations implied by the entity schema (e.g., alternative date/number readouts, spelling, honorifics); and (3) iteratively validate candidates using the same value--transcript consistency check, discarding patterns that frequently produce non-recoverable values. For new languages, the same structure applies: general LVPs are translated/parameterized, and entity-specific LVPs are adapted to locale-specific conventions (e.g., date order, numeral grouping, name particles).

% \begin{figure}[!h]
%   \centering
%   \includegraphics[width=\linewidth]{images/Similarity_score.pdf}
%   \caption{Framework to measure the similarity between \textsc{LingVarBench} generated transcripts and real phone call transcripts}
%   \label{fig:Similarity_score}
% \end{figure}

\label{sec:Appendix_LinguisticVariationTypes}
% --- Table 1: General ---
\begin{table*}[h]
\centering
\footnotesize
\renewcommand{\arraystretch}{1.2}
\begin{tabularx}{\linewidth}{l X}
\toprule
\textbf{Type} & \textbf{Instruction and Example} \\
\midrule
filler\_words & Include filler words like ``um'', ``uh'', ``you know''. Example: \texttt{um, it's one two three four five} \\
hesitation & Include hesitations and pauses. Example: \texttt{it's... one... two... three...} \\
correction & Include self-corrections. Example: \texttt{one two three... no wait, four five} \\
repetition & Repeat parts for emphasis. Example: \texttt{one two three, one two three, four five} \\
pause & Insert natural pauses. Example: \texttt{one two, pause, three four five} \\
formal & Use formal, precise language. Example: \texttt{the number is one two three four five} \\
casual & Use relaxed language. Example: \texttt{it's one two three four five} \\
polite & Use polite language. Example: \texttt{please, it's one two three four five} \\
confident & Sound confident. Example: \texttt{definitely one two three four five} \\
uncertain & Sound unsure. Example: \texttt{I think it's one two three four five} \\
rushed & Speak quickly. Example: \texttt{onetwothreefourfive} \\
careful & Speak slowly and carefully. Example: \texttt{carefully, one two three four five} \\
confirmation & Ask for confirmation. Example: \texttt{one two three four five, is that right?} \\
clarification & Clarify the answer. Example: \texttt{one two three four five, does that make sense?} \\
direct and simple & Be direct and simple. Example: \texttt{one two three four five} \\
brief\_confirmation & Use brief confirmation. Example: \texttt{yes, one two three four five} \\
concise\_confirmation & Use concise confirmation. Example: \texttt{confirmed, one two three four five} \\
\bottomrule
\end{tabularx}
\caption{linguistic verbalization patterns: \textbf{General} category.}
\label{tab:variation-types-general}
\end{table*}
% --- Table 2: ZIP Code Specific ---
\begin{table*}[!ht]
\centering
\footnotesize
\renewcommand{\arraystretch}{1.2}
\begin{tabularx}{\linewidth}{l X}
\toprule
\textbf{Type} & \textbf{Instruction and Example} \\
\midrule
digit\_by\_digit & Say each digit separately. Example: \texttt{one two three four five} \\
grouped\_two & Group digits in twos. Example: \texttt{twelve thirty-four five} \\
grouped\_three & Group digits in threes. Example: \texttt{one twenty-three forty-five} \\
hundred & Use ``hundred''. Example: \texttt{three hundred two five} \\
mixed\_grouping & Use mixed digit groupings. Example: \texttt{twelve three four five} \\
spoken\_number\_split & Split number words into digits. Example: \texttt{thirty two five eight} \\
reversed & Say digits in reverse. Example: \texttt{five four three two one} \\
with\_pause & Add pauses. Example: \texttt{one two... three four... five} \\
with\_repetition & Repeat groups. Example: \texttt{one two, one two, three four five} \\
with\_correction & Self-correct. Example: \texttt{one two three... no wait, four five} \\
with\_hesitation & Add hesitation. Example: \texttt{one... two... three... four... five} \\
with\_filler & Use filler words. Example: \texttt{um, one two three, you know, four five} \\
formal & Formal phrasing. Example: \texttt{the digits are one two three four five} \\
casual & Casual phrasing. Example: \texttt{yeah, it's one two three four five} \\
polite & Polite phrasing. Example: \texttt{please, it's one two three four five} \\
confident & Confident tone. Example: \texttt{definitely one two three four five} \\
uncertain & Uncertain tone. Example: \texttt{I think it's one two three four five} \\
spelled\_out & Spell digits with hyphens. Example: \texttt{one-two-three-four-five} \\
\bottomrule
\end{tabularx}
\caption{linguistic verbalization patterns: \textbf{ZIP code}-specific category.}
\label{tab:variation-types-zip}
\end{table*}

% --- Table 3: DOB Specific ---
\begin{table*}[h]
\centering
\footnotesize
\renewcommand{\arraystretch}{1.2}
\begin{tabularx}{\linewidth}{l X}
\toprule
\textbf{Type} & \textbf{Instruction and Example} \\
\midrule
date\_as\_4\_digits & 4-digit format. Example: \texttt{1267 → 01-02-1967} \\
spoken\_date\_4\_digits & Spoken version of 4-digit. Example: \texttt{one two six seven → 01-02-1967} \\
date\_as\_5\_digits & 5-digit format. Example: \texttt{32584 → 03-25-1984} \\
spoken\_date\_5\_digits & Spoken version of 5-digit. Example: \texttt{five one seven eight two → 05-17-1982} \\
date\_as\_6\_digits & 6-digit format MMDDYY. Example: \texttt{120285 → 12-02-1985} \\
spoken\_date\_6\_digits & Spoken 6-digit format. Example: \texttt{one two zero two eight five → 12-02-1985} \\
date\_as\_8\_digits & Full 8-digit date. Example: \texttt{12021947 → 12-02-1947} \\
spoken\_date\_8\_digits & Spoken 8-digit format. Example: \texttt{one two zero two one nine four seven → 12-02-1947} \\
spoken\_month\_day\_year & Natural spoken format. Example: \texttt{January second, nineteen ninety} \\
mixed\_spoken\_and\_digits & Mixed formats. Example: \texttt{January zero two, nineteen ninety} \\
filler\_or\_correction & Includes filler or correction. Example: \texttt{uh, zero one zero two one nine nine zero} \\
casual\_or\_polite\_digits & Casual/polite phrasing. Example: \texttt{please, one five, eighty five} \\
\bottomrule
\end{tabularx}
\caption{linguistic verbalization patterns, \textbf{Date of Birth (DOB)}-specific category.}
\label{tab:variation-types-dob}
\end{table*}

% --- Table 4: Name Specific ---
\begin{table*}[h]
\centering
\footnotesize
\renewcommand{\arraystretch}{1.2}
\begin{tabularx}{\linewidth}{l X}
\toprule
\textbf{Type} & \textbf{Instruction and Example} \\
\midrule
name\_with\_last & Full name. Example: \texttt{John Smith → John} \\
name\_with\_prefix & Prefix + name. Example: \texttt{My name is John Smith → John} \\
name\_reverse\_order & Last name first. Example: \texttt{Smith, John → John} \\
name\_with\_title & Name with title. Example: \texttt{Mr. John Smith → John} \\
name\_with\_middle & Name with middle. Example: \texttt{John Michael Smith → John} \\
name\_with\_suffix & Name with suffix. Example: \texttt{John Smith Jr. → John} \\
name\_with\_initials & Initials format. Example: \texttt{J. M. Smith → John} \\
name\_with\_correction & Correction. Example: \texttt{James—no, I mean John Smith → John} \\
name\_partial\_spelling & Partial spelling. Example: \texttt{John, that’s J-O-H-N Smith → John} \\
name\_with\_apostrophe & Apostrophe in last name. Example: \texttt{O'Connor, John → John} \\
name\_hyphenated & Hyphenated last name. Example: \texttt{John Smith-Jones → John} \\
nickname & Nickname. Example: \texttt{Johnny → John} \\
\bottomrule
\end{tabularx}
\caption{linguistic verbalization patterns: \textbf{Name}-specific category.}
\label{tab:variation-types-name}
\end{table*}

% --- Table: Pain Rating Specific ---
\begin{table*}[h]
\centering
\footnotesize
\renewcommand{\arraystretch}{1.2}
\begin{tabularx}{\linewidth}{l X}
\toprule
\textbf{Type} & \textbf{Instruction and Example} \\
\midrule
pain\_scale & Uses pain scale numbers or descriptive pain levels. Example: \texttt{seven, three out of ten, about a five} \\
comparative\_language & Uses comparative language when describing pain rating. Example: \texttt{worse than last time, about eight, not as bad, maybe four} \\
symptom\_description & Describes pain symptoms along with the rating. Example: \texttt{it's a seven, sharp and throbbing, about five, dull ache} \\
health\_assessment & Includes health context with pain rating. Example: \texttt{considering my condition, I'd say seven, for someone my age, probably a six} \\
confident & Shows confidence when stating pain rating. Example: \texttt{definitely seven, it's absolutely an eight, for sure five} \\
hesitant & Shows hesitation when stating pain rating. Example: \texttt{um... maybe seven?, I guess... five?, well... probably six} \\
uncertain & Expresses uncertainty about pain rating. Example: \texttt{I'm not sure, maybe seven, I think it's around five, probably six, I guess} \\
formal & Uses formal language when stating pain rating. Example: \texttt{I would rate it at seven, the pain level is approximately five, my pain rating is eight} \\
casual & Uses casual language when stating pain rating. Example: \texttt{yeah, it's like a seven, oh, maybe five, I'd say six} \\
polite & Uses polite language when stating pain rating. Example: \texttt{I would say seven, please, it's five, thank you for asking, about six, if I may} \\
filler\_words & Includes filler words when stating pain rating. Example: \texttt{um, it's like, you know, seven, well, uh, about five, so, like, maybe six} \\
hesitation & Includes pauses and false starts when stating pain rating. Example: \texttt{it's... seven, I would say... five, um... eight} \\
correction & Includes self-corrections when stating pain rating. Example: \texttt{six... no wait, seven, I said five, but actually six, eight... or maybe seven} \\
repetition & Repeats the pain rating for emphasis. Example: \texttt{seven, seven, it's five, five out of ten, eight, definitely eight} \\
pause & Includes natural pauses when stating pain rating. Example: \texttt{it's... about seven, let me think... five, I'd say... six} \\
thoughtful & Speaks slowly and deliberately when stating pain rating. Example: \texttt{let me think carefully... seven, well... I would say... five, hmm... probably six} \\
confused & Shows confusion about how to rate pain. Example: \texttt{I'm not sure how to rate it... seven?, is five a lot? I'll say five, what does seven mean exactly?} \\
frustrated & Shows frustration when stating pain rating. Example: \texttt{ugh, it's like a seven, I already said eight, for the last time, five} \\
rushed & Speaks quickly when stating pain rating. Example: \texttt{seven quickly, just five, six let's move on} \\
\bottomrule
\end{tabularx}
\caption{linguistic verbalization patterns: \textbf{Pain Rating}-specific category.}
\label{tab:variation-types-pain-rating}
\end{table*}

% --- Table 5: Respiratory Issues Specific ---
\begin{table*}[h]
\centering
\footnotesize
\renewcommand{\arraystretch}{1.2}
\begin{tabularx}{\linewidth}{l X}
\toprule
\textbf{Type} & \textbf{Instruction and Example} \\
\midrule
condition\_specific & Specify the exact condition (asthma, COPD, or both). Example: \texttt{Yes, I have asthma} \\
severity\_level & Mention severity or frequency of the condition. Example: \texttt{Yes, severe asthma} \\
treatment\_mention & Reference treatment while answering. Example: \texttt{Yes, I use an inhaler for asthma} \\
doctor\_reference & Reference doctor or medical diagnosis. Example: \texttt{My doctor diagnosed me with asthma} \\
medical\_terminology & Use formal medical terminology. Example: \texttt{I have chronic obstructive pulmonary disease} \\
health\_concern & Express health concerns or impact. Example: \texttt{Yes, asthma, it affects my breathing} \\
confident & Answer with confidence. Example: \texttt{Yes, definitely have asthma} \\
hesitant & Answer with hesitation. Example: \texttt{Well... I think I have asthma} \\
uncertain & Express uncertainty about the condition. Example: \texttt{I think it's asthma} \\
formal & Use formal language. Example: \texttt{Yes, I have been diagnosed with asthma} \\
casual & Use casual language. Example: \texttt{Yeah, got asthma} \\
polite & Use polite, respectful language. Example: \texttt{Yes, I do have asthma, thank you for asking} \\
filler\_words & Include filler words. Example: \texttt{Um, yes, I have asthma} \\
hesitation & Include hesitations and pauses. Example: \texttt{Yes... I have... asthma} \\
correction & Include self-corrections. Example: \texttt{Asthma... no wait, COPD} \\
repetition & Repeat parts for emphasis. Example: \texttt{Yes, yes, I have asthma} \\
pause & Include natural pauses. Example: \texttt{Yes, [pause] asthma} \\
thoughtful & Sound thoughtful or reflective. Example: \texttt{Let me think... yes, asthma} \\
confused & Sound confused about the question. Example: \texttt{Wait, asthma or COPD? I have asthma} \\
frustrated & Express frustration. Example: \texttt{Yes, I already said, asthma!} \\
rushed & Answer quickly or hurriedly. Example: \texttt{Yeah-asthma} \\
\bottomrule
\end{tabularx}
\caption{linguistic verbalization patterns: \textbf{Respiratory Issues}-specific category.}
\label{tab:variation-types-respiratory-issues}
\end{table*}

\end{document}